\documentclass[final]{cvpr}

\usepackage{times}
\usepackage{epsfig}
\usepackage{graphicx}
\usepackage{amsmath}
\usepackage{amssymb}
\usepackage{multicol}
\usepackage{latexsym}
\usepackage{subfigure}
\usepackage{placeins}
 
\usepackage{xcolor}
\usepackage{color}
\usepackage{soul}

\usepackage{tabularx}
\usepackage{multirow}
\usepackage{float}

\usepackage[pagebackref=true,breaklinks=true,colorlinks,bookmarks=false]{hyperref}

\begin{document}

\title{Detail Preserving Residual Feature Pyramid Modules for Optical Flow}

\author{Libo Long ~~~~~~~~~~~~~~~~~~~~~~~~~~~
Jochen Lang\\
EECS, University of Ottawa\\
{\tt\small llong014@uottawa.ca} ~~~~~~~~~~~~~
{\tt\small jlang@uottawa.ca}\\
}

\maketitle

\begin{abstract}
Feature pyramids and iterative refinement have recently led to great progress in  optical flow estimation. However, downsampling in feature pyramids can cause blending of foreground objects with the background, which will mislead subsequent decisions in the iterative processing. The results are missing details especially in the flow of thin and of small structures. We propose a novel Residual Feature Pyramid Module (RFPM) which retains important details in the feature map without changing the overall iterative refinement design of the optical flow estimation. RFPM incorporates a residual structure between multiple feature pyramids into a downsampling module that corrects the blending of objects across boundaries. We demonstrate how to integrate our module with two state-of-the-art iterative refinement architectures. Results show that our RFPM visibly reduces flow errors and improves state-of-art performance in the clean pass of Sintel, and is one of the top-performing methods in KITTI. According to the particular modular structure of RFPM, we introduce a special transfer learning approach that can dramatically decrease the training time compared to a typical full optical flow training schedule on multiple datasets.
\end{abstract}

\section{Introduction}

\begin{figure}[htb]

\subfigure[Boundary blur on a multi-scale image. In the high resolution image (top), we can easily distinguish the outline of individual apples, while in the low resolution image (bottom), it is difficult to recognize individual apples due to their blurred boundaries.]{%
  \includegraphics[clip,width=\columnwidth]{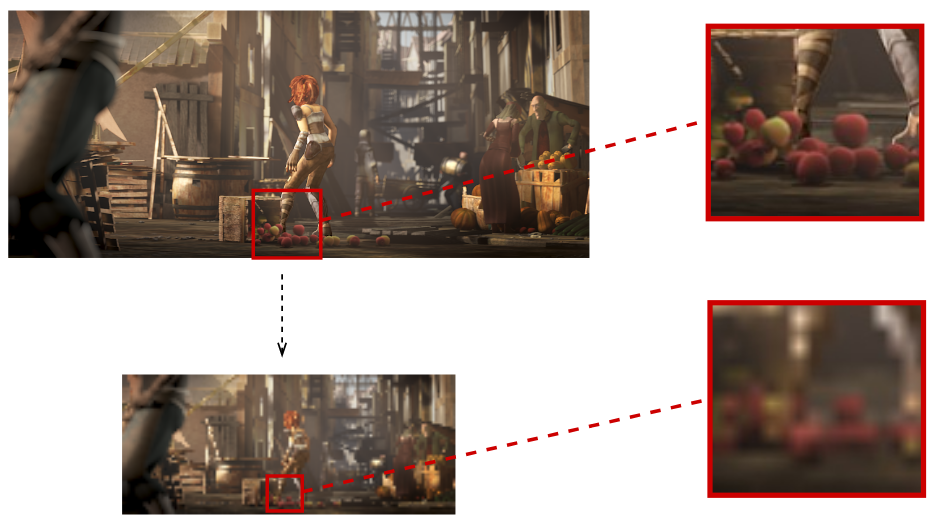}%
}

\subfigure[Boundary blur on multi-scale feature map. The top image is the feature map at Level 2 of the pyramid where we can see the approximate boundaries of each apple. The bottom image is the feature map at Level 5 where the features in the highlighted area are not discernible.]{%
  \includegraphics[clip,width=\columnwidth]{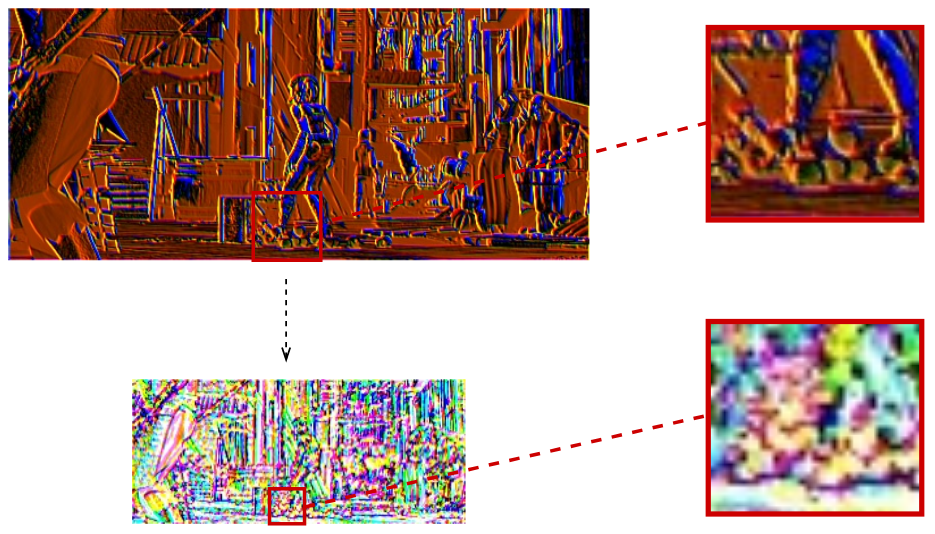}%
}

\caption{Failure cases in current pyramid architectures. (a) Low resolution images blur object boundaries. (b) The feature map exhibits the same problem due to the down-sampling in the convolution layer.\label{fig:boundaryBlur}}

\end{figure}

Optical flow estimation is a key problem in computer vision and a fundamental building block of many high-level computer vision application~\cite{7298925,NIPS2014_5353,Lai-ECCV-2018}. Recent optical flow estimation has greatly benefited from learning-based CNN architectures. Flownet~\cite{DFIB15} is the first CNN which directly predicts the optical flow with an encoder-decoder architecture. PWC-Net~\cite{Sun2018PWC-Net} and LiteFlowNet~\cite{hui18liteflownet} proposed an iterative refinement design. The four fundamental stages in iterative refinement are: first feature maps at different levels of resolution are extracted; second, correlation is used to calculate a cost volume; third, intermediate optical flow is predicted based on the cost volume and on the previous optical flow; and finally, the previous three steps are repeated in an iterative refinement loop. This architecture has been shown to effectively reduce error in large displacement and it has been used as a design in many recent approaches~\cite{DBLP:journals/corr/abs-1904-09117, NEURIPS2019_bbf94b34, Hur:2020:SSM,GLUNet_Truong_2020,hu2016cpm}. However, we observe that the iterative refinement design is not without major drawbacks which limits optical flow estimation to improve further. Fig.~\ref{fig:boundaryBlur} shows that the low-resolution image, as well as the low-resolution feature map, exhibit blending across boundaries of foreground objects (the apples in Fig.~\ref{fig:boundaryBlur}) and the background because of down-sampling. This can mislead the flow estimation to incorrectly consider foreground and background as one object and hence predict the optical flow incorrectly for both, foreground and background. Because of the iterative design, the erroneous estimate will be amplified in the following iterations.

We introduce a new pyramid module, Recurrent Feature Pyramids Module (RFPM), to address the loss of detail in feature pyramids. We demonstrate how RFPM reduces error in optical flow in Sintel and KITTI, even when integrated into top-performing optical flow methods such as RAFT~\cite{2020arXiv200312039T} and IRR-PWC~\cite{Hur:2019:IRR}. Our experimental results shows that RFPM-RAFT achieves state-of-the-art performance on MPI Sintel~\cite{Butler:ECCV:2012} (Clean pass), and KITTI 2015~\cite{Menze2018JPRS} benchmarks (two-frame).

We decide to focus on the feature pyramid because we do not want to change the iterative refinement architecture as it has been shown to improve optical flow results, especially for large flow, and because the pyramid module is fully compatible with most optical flow estimation methods or high-level applications. In order to reduce the effort in re-training a complete optical flow method, we also propose a new transfer learning strategy which directly adds RFPM to a trained optical flow method and thereby significantly increasing training progress. 

Our main contributions are: First, we show that the traditional pyramid architecture is flawed in optical flow estimation. We present a new pyramid architecture, RFPM, which reduces the error at motion boundaries. We then propose a new strategy of transfer learning for RFPM. Next, we will review related optical flow methods, strategies to avoid blur across motion boundaries and the use of feature pyramids.

\section{Related work} 

\noindent\textbf{Optical Flow Estimation.} In optical flow the use of multi-resolution representations goes back to Lucas and Kanade~\cite{lucas1981iterative} who used bandpass-filtering to register features across scales. Bouguet~\cite{bouguet2001pyramidal} reported a popular implementation using image pyramids. We refer the reader to the work of Sun et al.~\cite{sun2014quantitative} and their analysis of classic optical flow methods derived from Horn and Schunck~\cite{horn1981determining}. FlowNet~\cite{DFIB15} is the first end-to-end trainable CNN network based on a U-Net~\cite{RFB15a} architecture. The model was trained on a synthetic dataset. They propose two basic architectures but the  correlation layer of FlowNetC has become a key component in modern architectures. Next, Ilg et al.~\cite{IMKDB17} stack several models of FlowNet into a large system. With a regular training schedule, FlowNet2 achieves significant improvements in the Sintel and KITTI benchmarks. SpyNet~\cite{Ranjan_CVPR_2017} introduces a light-weight network by using the feature pyramid and warping within the image pyramid. Sun et al.~\cite{Sun2018PWC-Net} create PWC-Net that uses an architecture that utilizes a feature pyramid, warping, and a cost volume, which forms the cornerstone of many follow-up works~\cite{DBLP:journals/corr/abs-1904-09117,Hur:2019:IRR,8658399}. Recently, IRR~\cite{Hur:2019:IRR} and RAFT~\cite{2020arXiv200312039T} apply an iterative architecture, which use a fixed CNN component as an unit in the iteration. This architecture achieves remarkable performance with fewer parameters compared to most conventional CNN architectures.

\noindent\textbf{Boundary Blur.} Blur across optical flow boundariers is known to be a difficult problem. Many methods use segmentation to prevent blur. Earlier works~\cite{1544871,541407} segment the image into regions using shape or color, then estimating motion by matching the regions. In recent CNN based methods, Sevilla et al.~\cite{7780791} and Hur et al.~\cite{hur2016joint} add a pre-trained semantic segmentation neural network as an additional component to improve optical flow but these methods are not end-to-end trainable. SegFlow~\cite{DBLP:journals/corr/abs-1709-06750} uses an optical flow and an image segmentation model jointly, and construct communications between the two branches of the network. Different from previous work, our method does not use an extra semantic network instead we prevent boundary blur with the proposed pyramid architecture.

\noindent\textbf{Feature Pyramids.} Image pyramids as a multi-resolution representation are widely used in image processing, computer vision and computer graphics~\cite{article,adelson1984pyramid}. Pyramids of image features are often calculated based on multi-resolution images. This method is slow however, as hand-engineered features on each scale of the images need to be computed. Liu et al.~\cite{DBLP:journals/corr/LiuAESR15} use the convolution layer to predict multi-scale feature maps in their SSD to handle variedly sized objects. However, as noted by Fu et al.~\cite{DBLP:journals/corr/FuLRTB17}, SSD fails to detect small instances. Subsequently, Feature Pyramid Networks~\cite{DBLP:journals/corr/LinDGHHB16} are a top-down architecture with skip connections in the feature pyramid combining semantically-strong features at low resolution with semantically-weak features at high resolution. This architecture makes the network more robust for different scales of objects. Kong et al.~\cite{DBLP:journals/corr/abs-1808-07993} add global attention and local reconfiguration into a SSD-like design. In their work, they compare SSD and its variants showing that global attention enhances multi-scale representations with semantically strong information. The above methods address object detection, semantic segmentation etc.\ but do not consider optical flow. 

\begin{figure}[h]
\begin{center}
   \includegraphics[width=0.65\columnwidth]{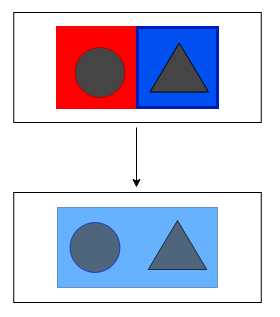}
\end{center}
   \caption{Illustration of flow error. The top shows two pixels of a high resolution image, red indicates the flow of the object is to the left, and blue indicates the flow of the object is to the right. The top high resolution image predicts the flow of the two objects correctly (circle towards left, triangle towards right). The low resolution image at the bottom demonstrates that fusing the two pixel produces an average flow prediction which is different from either correct flow value.\label{fig:tr}}
\end{figure}

Recent optical flow estimation methods \cite{Hur:2019:IRR,Sun2018PWC-Net,zhao2020maskflownet} use a shared weight pyramid to extract features. The main difference of pyramids in optical flow methods from image detection is that optical flow methods use iterative refinement while in object detection the result is directly predicted. 
To our knowledge, no prior work has focused on feature pyramids in optical flow in end-to-end trainable optical flow architectures; most of the optical flow methods still use an SSD-like architecture. However, the limitations of SSD for small objects are well known and the problem is exaggerated by the iterative refinement processing in optical flow. 

\begin{figure}[htb]
\begin{center}
\subfigure[\textbf{Weighted Feature Downsampling (WFD)}]{%
  \includegraphics[clip,width=0.7\columnwidth]{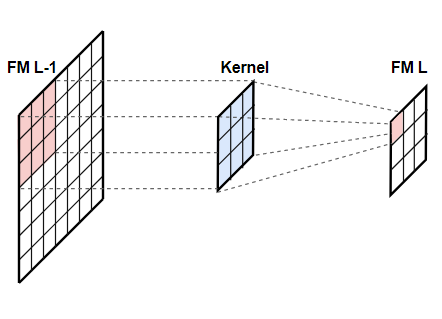}%
}


\subfigure[\textbf{Residual Feature Downsampling (RFD)}]{%
  \includegraphics[clip,width=0.7\columnwidth]{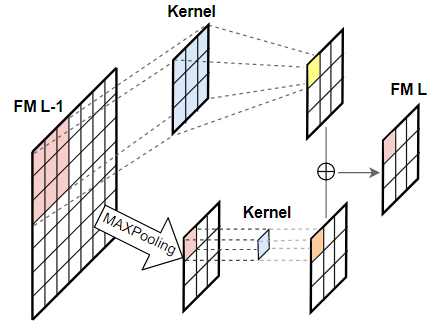}%
}
\end{center}

\caption{Downsampling modules. WFD is the original module used in CNN optical flow estimation~\cite{Sun2018PWC-Net,hui20liteflownet3,Hur:2019:IRR}. RFD combines WFD and Max Pooling (MP) with a residual-like design. \label{fig:3}}

\end{figure}

\section{Feature Map Analysis}
\label{seg:featureMap}

Recent iterative refinement optical flow methods calculate feature pyramids and then use these multi-resolution maps in a correlation step to find a cost volume. We will consider first the currently used approach which we refer to as Weighted Feature Downsampling (WFD) before deriving our alternative approach. 

Given a pair of RGB images, $I_t$, $I_{t+1}$, a feature pyramid generates L-level bottom-up pathways, the bottom level is the original input image, i.e., $\mathcal{F}^0_t = I_t$. Each feature map $\mathcal{F}^l_t$ is generated typically by a $3 \times 3$ convolutional filter with a stride of two pixels with respect to $\mathcal{F}^{l-1}_t$. The architecture contains 6 levels of convolutional blocks in recent works~\cite{Sun2018PWC-Net,Hur:2019:IRR} while using the last four feature map pairs \{($\mathcal{F}^3_t,\mathcal{F}^3_{t+1}
$), ($\mathcal{F}^4_t,\mathcal{F}^4_{t+1}$), ($\mathcal{F}^5_t,\mathcal{F}^5_{t+1}$), ($\mathcal{F}^6_t,\mathcal{F}^6_{t+1}$)\} to compute the cost volume. Therefore, at the l-th level, the cost volume can be formulated as 
\begin{equation}
V = c(\mathcal{F}^l_t,w(\mathcal{F}^l_{t+1})),
\label{eqn:costVolume}
\end{equation}
where $w$ represents the warping operation that uses bi-linear interpolation to warp $I_{t+1}$ with the optical flow to $I_{t}$. The correlation operation $c$ computes the similarity between two feature maps by an element-wise dot product. 

At each level $l$ of the feature pyramid, the output pixel value $x^l_{i,j}$ is a weighted sum of  pixel value of the previous level \begin{equation}
z^l_{i,j} = \sum_{x,y}z^{l-1}_{i+x,j+j}k_3(x,y),
\end{equation}
where i,j are the indices of the pixel, $k_3$ is the 3*3 convolution kernel, $-1 \leq x,y \leq 1$ (3*3 kernel). We refer to this as Weighted Feature Downsampling (WFD) in Fig. \ref{fig:3}(a).

We observe a major consequence of this architecture is blurring of motion boundaries. Fig.~\ref{fig:tr} illustrates an example: Two tiny objects move in different directions in the original images $I_t$ and $I_{t+1}$ after WFD through several convolution layers. The coarse resolution treats the two objects as a whole, and predicts the same or similar direction for both objects which is incorrect for both and contradicts a possibly correct predictions at the higher resolution. As a result the overall prediction will be in error.


\noindent\textbf{Residual Feature Downsampling (RFD).}
We propose to apply a joint feature extraction by WFD and Max Pooling (MP) in a residual design~\cite{DBLP:journals/corr/VeitWB16}. RFD learns to enhance edge areas because MP is expected to keep higher weight pixels for the next level. Although MP itself might not keep all significant information, it can guide WFD towards the extra information through the residual structure. To this end, we introduce a Residual Feature Downsampling (RFD) pyramid based on the integration of WFD and MP. The RFD extracts features with an addition operation from both WFD and MP (see Fig.~\ref{fig:3}(b)). The pixel value at level $l$ is then
\begin{equation}
z^l_{i,j} = \sum_{x,y} M(z)^{l-1}_{i+x,j+j}k_1(x,y)+z^{l-1}_{i+x,j+j}k_3(x,y)
\end{equation}
Our RFPM includes RFD but it also incorporates multiple different feature pyramids, including pyramids calculated with standard WFD. We also provide additionally pathways for the feature in form of repair masks that act at a chosen level of different pyramids.


\begin{figure}[bt]
\begin{center}
   \includegraphics[width=1\linewidth]{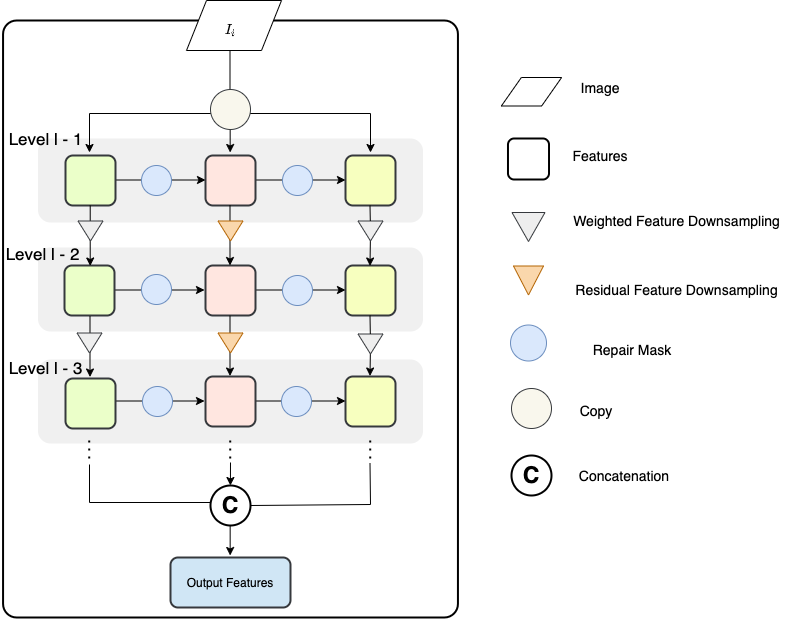}
\end{center}
   \caption{Architecture of RFPM. Feature maps are extracted with three different pyramids. \label{fig:overall}}
\end{figure}

\begin{figure}[bt]
\begin{center}
   \includegraphics[width=1\linewidth]{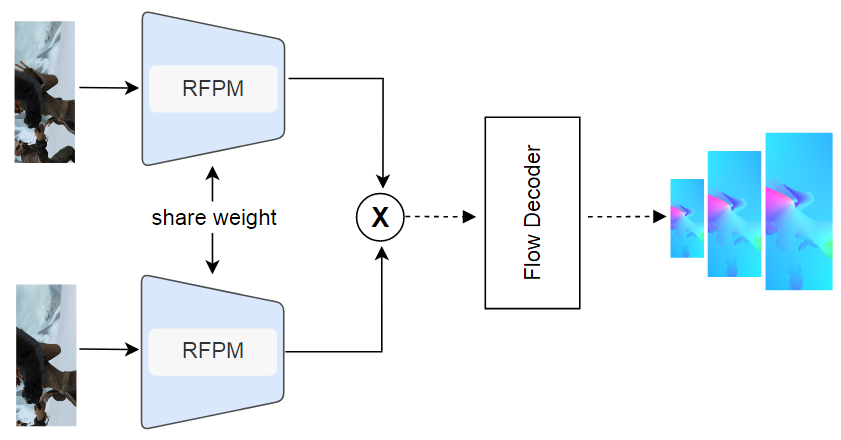}
\end{center}
   \caption{The network extracts a pair of feature maps by a share weight RFPM and feeds them into a correlation layer.}
\end{figure}

\subsection{Repair Mask}
We introduce a learnable Repair Mask (RM) to repair feature maps after downsampling. The RM contains two parts: a multiplicative attention function $\mathcal{A}$ which output a attention mask of shape (B, 1, H, W), and an additive bias function $\mathcal{M}$ which output a bias mask of shape (B, C, H, W). Assumig two pyramids (and using broadcasting of $\mathcal{A}$), then the feature map of the right pyramid at level $l+1$ takes the map of the left pyramid into account as in
\begin{equation}
\mathcal{F}^{l+1}_{right} = Conv(\mathcal{F}^{l}_{right}*\mathcal{A}(\mathcal{F}^{l+1}_{left})+\mathcal{M}(\mathcal{F}^{l+1}_{left}))
\end{equation}

\subsection{Module Structure}

\begin{figure*}[t]
\begin{center}
   \includegraphics[width=1\linewidth]{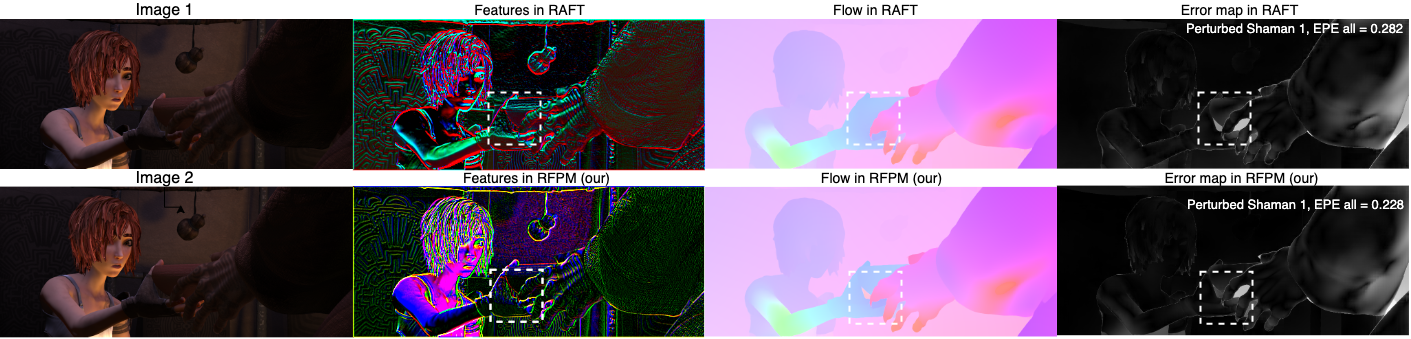}
   
   \includegraphics[width=1\linewidth]{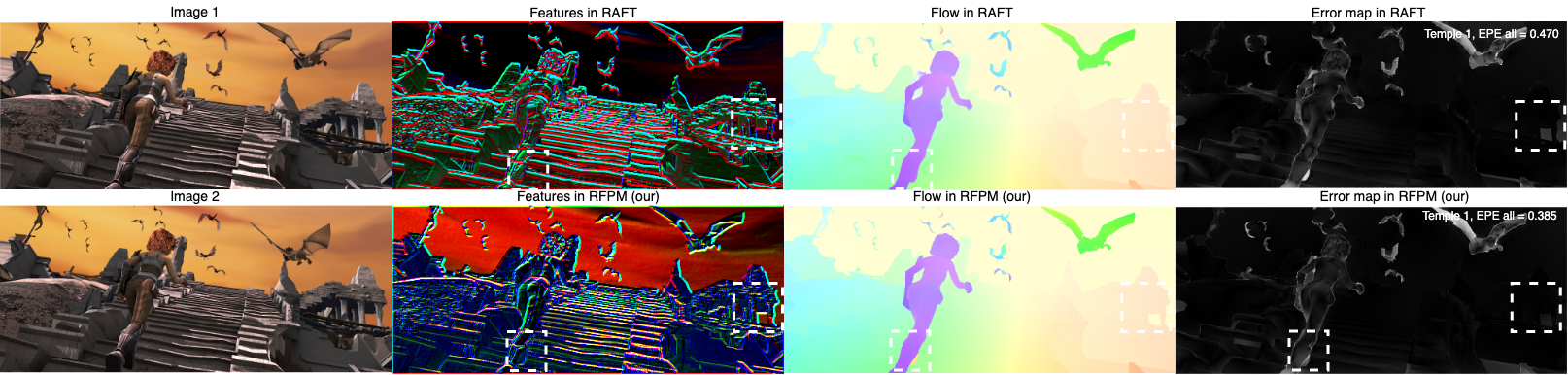}
\end{center}
\caption{Visualizing the relation between feature map and predicted flow. Failure areas in RAFT are indicated with white squares. These areas have blurred edges in the feature pyramid. In our RFPM, the flow result correctly predicts the motion boundary because of the improved edges in the feature pyramid. RFPM significantly reduces the EPE errors.\label{fig:flow}}
\end{figure*}

\begin{figure*}[h]
\begin{center}
   \includegraphics[width=1\linewidth]{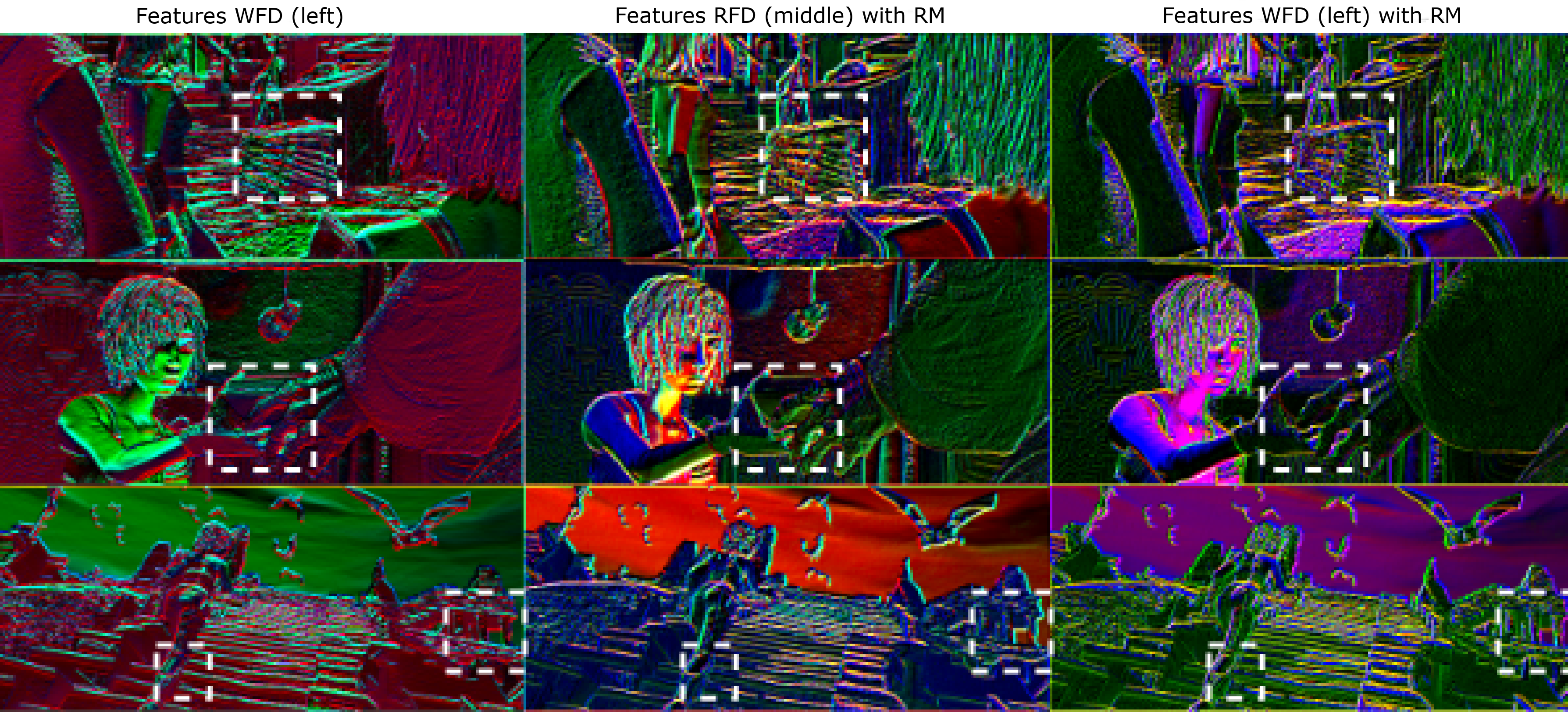}
\end{center}
\caption{Visualization of feature pyramids. Comparison of the learned features from feft pyramid using WFD, middle pyramid using RFD with repair mask (RM), and right pyramid using WFD with repair mask (RM). \label{fig:features}}
\end{figure*}

The complete Recurrent Feature Pyramid Module with Repair Masks (RFPM) 
is constructed as follows (see Fig.~\ref{fig:overall}):

(a) Multi-kernel top-down pathway: Given a pair of images, features are extracted from the top to bottom via three pathways, the left and right pyramid are constructed with WFD. We call the left pyramid the base pyramid. The middle pyramid uses RFD. When integrating RFPM into different optical flow estimators, the actual pyramid architecture needs to be slightly different. We always keep the original architecture as base pyramid. This design of RFPM is also beneficial for transfer learning which we will describe in Section~\ref{sec:training}.

(b) Repair left-right pathway: The feature map of the base pyramid extracts a repair mask, the mask passes to the middle pyramid to repair missing information due to maximum pooling. The middle pyramid in turn extracts a repair mask to restore the right pyramid as well. Note, we do not add a repair mask at each level. In practice, we found adding repair masks only on some levels at the bottom of the pyramid is sufficient to improve performance. 

Therefore, at level $l$, the cost volume is defined as in Eqn.~\ref{eqn:costVolume} with the feature maps:
\begin{eqnarray}
\mathcal{F}^l_{t} & = & [\mathcal{F}^l_{left,t},\mathcal{F}^l_{mid,t},\mathcal{F}^l_{right,t}] \\
%
\mathcal{F}^l_{t+1} & = & [\mathcal{F}^l_{left,t+1},\mathcal{F}^l_{mid,t+1},\mathcal{F}^l_{right,t+1}]
\end{eqnarray}


Next, we will discuss how to integrate RFPM with two state-of-the-art optical flow methods: IRR-PWC~\cite{Hur:2019:IRR} and RAFT~\cite{2020arXiv200312039T}.

\subsection{RFPM with IRR-PWC}
RFPM-IRR-PWC consists of a 6-level shared-weight base pyramid with 16, 32, 64, 96, 128 and 196 feature channels, respectively. Each level from 6 downto 2 makes predictions in a coarse-to-fine manner. The first predictions at level 6 are at $1\over 64$ and the final predictions at level 2 are at $1\over4$ of the resolution of the original image in width and height. The feature channels of the 
pyramids are 16, 32, 64, 88, 112 and 136, respectively. Fig.~\ref{fig:overall} illustrates the pyramids with repair masks generated at level 1 to 3. The base pyramid provides mask $\mathcal{A}(\mathcal{F}^{l}_{left})$ and $\mathcal{M}(\mathcal{F}^{l}_{left})$, which feed into the same level of the middle pyramid. This avoids unnecessary down- and upsampling. The middle pyramid generates masks $\mathcal{A}(\mathcal{F}^{l}_{mid})$ and $\mathcal{M}(\mathcal{F}^{l}_{mid})$, which feed into the same level of the right pyramid. Due to the bilateral refinement of flow and occlusion structures, feature maps of all three pyramids feed into the correlation layer. In contrast, only the feature map of the base pyramid is used to predict occlusions and the context flow.

\subsection{RFPM with RAFT}
RAFT contains a three-level feature pyramid, each level consists of two residual blocks, at $1\over2$, $1\over4$, and $1\over8$ resolutions of the original image in width and height, respectively. Instead of a coarse-to-fine manner, RAFT builds a multi-scale 4D correlation volume by all pairs of pixels in the feature map of the last level. RFPM-RAFT uses a similar modification than in RFPM-IRR-PWC, except RFPM-RAFT uses the three-level pyramid architecture of RAFT and the repair masks are generated at level 1 and level 2.

\section{Experimental Evaluation}

We first detail the training for our implementation of RFPM-IRR-PWC and RFPM-RAFT before presenting results on MPI Sintel~\cite{Butler:ECCV:2012} and KITTI-2015~\cite{Menze2018JPRS}. We present ablation studies for the configurations of the downsampling approaches as well as the number and level of repair masks in RFPM. We then discuss our reduced effort training approach for our module based on transfer learning and data augmentation.

\subsection{Training and Implementation}
\label{ref:training}

We follow the training configurations of IRR-PWC \cite{Hur:2019:IRR} and RAFT \cite{2020arXiv200312039T} for a fair comparison. We first train RFPM-IRR-PWC on the FlyingChairs\_OCC~\cite{DFIB15} dataset (learning rate schedule schedule S$_{short}$) and then fine tune on FlyingThings3D~\cite{MIFDB16} (half schedule learning rate of S$_{short}$). When fine-tuning on Sintel and KITTI, we use a mini-batch size of 4 with the cyclic learning rate proposed by PWC-Net+~\cite{Sun2018:Model:Training:Flow}. IRR-PWC uses fine-tuning on KITTI based on the checkpoints of FlyingThings3D.
We train RFPM-RAFT on FlyingChairs for 200k iterations with a batch size of 6, and on FlyingThing3D for 200k iterations with a batch size of 3. We fine-tune on Sintel by combining data from Sintel~\cite{Butler:ECCV:2012}, KITTI-2015~\cite{Menze2018JPRS} and HD1K~\cite{7789500} for 200k iterations with a batch size of 3 (the same setting as described in PWC-Net+~\cite{Sun2018:Model:Training:Flow}).
Then, based on the checkpoints of Sintel, we train for 100k iterations with a batch size of 3 (we use half of the batch size and double the iterations in comparison with RAFT because of memory limitations). RFPM is implemented in PyTorch~\cite{NEURIPS2019_9015} and our experiments use either a single Nvidia 2080Ti GPU with 11 GB of memory, or a Nvidia RTX2070 GPU with 8GB of memory. Some of the ablation study were done with Google Colab. 

\begin{table*}[t!]
\begin{tabularx}{\textwidth}{c @{\extracolsep{\fill}} llllllllll}
\hline\hline
\multicolumn{2}{c}{\multirow{2}{*}{Method}} & \multicolumn{2}{c}{Sintel (train)} & \multicolumn{2}{c}{KITTI-15 (train)} & \multicolumn{2}{c}{Sintel (test)} & \multicolumn{2}{c}{KITTI-15 (test)}\\

\cline{3-4} \cline{5-6}  \cline{7-8} \cline{9-10}
     &   &  Clean   & Final   & F1-epe& F1-all& Clean &Final &\multicolumn{2}{c}{F1-all(\%)}                                        
	
	\\    \hline

\multicolumn{2}{l}{SPyNet\cite{Ranjan_CVPR_2017}} & (3.17) & (4.32) & - & - & 6.64 & 8.36 & \multicolumn{2}{c}{35.07}   \\	
\multicolumn{2}{l}{FlowNet2\cite{IMKDB17}} & (1.45) & (2.19) & (2.36) & (8.88) & 4.16 & 5.74 & \multicolumn{2}{c}{10.41}  \\  
\multicolumn{2}{l}{FlowNet3\cite{ilg2018occlusions}} & (1.47) & (2.12) & (1.79) & - & 4.35 & 5.67 & \multicolumn{2}{c}{8.60} \\
\multicolumn{2}{l}{SelFlow\cite{DBLP:journals/corr/abs-1904-09117}} & (1.68) & (1.77) & (1.18) & - & 3.75 & 4.26 & \multicolumn{2}{c}{8.42}   \\
\multicolumn{2}{l}{PWC-Net\cite{Sun2018PWC-Net}} & (2.02) & (2.08) & (2.16) & (9.80) & 4.39 & 5.04 & \multicolumn{2}{c}{9.60}   \\
\multicolumn{2}{l}{PWC-Net+\cite{Sun2018:Model:Training:Flow}} & (1.71) & (2.34) & (1.47) & (7.59) & 3.45 & 4.60 & \multicolumn{2}{c}{7.72}   \\
\multicolumn{2}{l}{IRR-PWC\cite{Hur:2019:IRR}} & (1.92) & (2.51) & (1.63) & (5.32) & 3.84 & 4.58 & \multicolumn{2}{c}{7.65}   \\
\multicolumn{2}{l}{LiteFlowNet2\cite{hui20liteflownet2}} & (1.30) & (1.62) & (1.33) & (4.32) & 3.48 & 4.69 & \multicolumn{2}{c}{7.62}   \\
\multicolumn{2}{l}{LiteFlowNet3\cite{hui20liteflownet3}} & (1.32) & (1.76) & (1.26) & (3.82) & 2.99 & 4.45 & \multicolumn{2}{c}{7.34}   \\
\multicolumn{2}{l}{HD3\cite{Yin_2019_CVPR}} & (1.70) & (1.17) & (1.31) & (4.10) & 4.79 & 4.67 & \multicolumn{2}{c}{6.55}   \\
\multicolumn{2}{l}{VCN\cite{NEURIPS2019_bbf94b34}} & (1.66) & (2.24) & (1.16) & (4.10) & 2.81 & 4.40 & \multicolumn{2}{c}{6.30}   \\
\multicolumn{2}{l}{MaskFlowNet\cite{zhao2020maskflownet}} & - & - & - & -& 2.52 & 4.17 & \multicolumn{2}{c}{6.10}  \\
\multicolumn{2}{l}{RAFT\cite{2020arXiv200312039T}} & (0.76) & (1.22) & (0.63) & (1.50) & 1.94 & 3.18 & \multicolumn{2}{c}{5.10}   \\
\multicolumn{2}{l}{RAFT(warm-start)\cite{2020arXiv200312039T}} & (0.77) & (1.27) & - & - & 1.61 & \textbf{2.86} & \multicolumn{2}{c}{-} \\

\multicolumn{2}{l}{RAFT-A*\cite{RAFT_A}} & - & - & - & - & 2.01 & 3.14 & \multicolumn{2}{c}{\textbf{4.78}} \\

   \hline
                            
\multicolumn{2}{l}{RFPM-IRR-PWC} & (1.66) & (2.43) & 1.48 & 5.17 & 3.63 & 4.52 & \multicolumn{2}{c}{7.49}\\ 
\multicolumn{2}{l}{RFPM-RAFT} & (\underline{0.61}) & (\underline{1.05}) & (\underline{0.60}) & (\underline{1.41}) & - & - & \multicolumn{2}{c}{4.79}   \\
\multicolumn{2}{l}{RFPM-RAFT(warm-start)} & (0.68) & (1.12) & - & - & \textbf{1.41} & 2.90 & \multicolumn{2}{c}{-} \\
\hline\hline
\end{tabularx}
\caption {\label{tab:table1} Results Comparison on Sintel and KITTI-15. The value in parentheses are the errors on the training dataset, the best training result is underlined, and the best testing result is shown in bold. Sintel performance is evaluated by average end-point error (AEPE) over all valid pixels. F1-all is the percentage of optical flow outliers over all valid pixels. * Please note that AutoFlow by Sun et al.\cite{RAFT_A} uses a different training dataset but does not change the basic RAFT architecture.} 
\end{table*}

\subsection{Results}
According to Table~\ref{tab:table1}, RFPM-RAFT outperforms all published optical flow methods on the MPI Sintel~\cite{Butler:ECCV:2012} clean pass and KITTI-2015~\cite{Menze2018JPRS}. Our methods achieve a 12\% higher accuracy on Sintel, and 6\% higher accuracy on KITTI compared with RAFT. Fig.~\ref{fig:flow} shows a visualization of test results from the Sintel website. To show the benefit of our methods, we compare RAFT with our  RFPM-RAFT in both feature map (second column) and output flow (third column). The dashed rectangle corresponds to the same zoomed-in areas as in the feature map and flow results. Comparing features in RAFT and in our proposed RFPM-RAFT, we clearly see that features in RAFT ignore important edges in the zoom-in, and the output flow thus failed to predict motion boundary in the corresponding area (dashed rectangle). In contrast, our RFPM-RAFT significantly reduces the error and better separates different moving objects and environments in these areas. Furthermore, Fig.~\ref{fig:features} visualizes all three feature maps in the same level of RFPM-RAFT, we can see that by adding the repair mask in the RFD (middle) and WFD (right), the edges in the feature map are clearer than in the original WFD (left). Therefore, we think the missing detection of edges in the feature maps is the main reason for a poor prediction of optical flow close to motion boundaries in RAFT. As can be seen, our method enhances the edges in the rectangle and succeeds to estimate the optical flow close to these motion boundaries.

\begin{table}[h!]
\centering
\begin{tabularx}{\linewidth}{ccccc}
\hline\hline
\multicolumn{2}{c}{\multirow{3}{*}{Settings}} & \multicolumn{3}{c}{Trained on Chairs} \\
 
\cline{3-5} 

     &   &  Chairs   & Sintel(clean)   & Sintel(final)\\
     &   &  Testing   & Training   & Training

	\\    \hline
	 \multicolumn{2}{l}{L/M}  &  0.79   & 2.29   & 4.41\\
	 \multicolumn{2}{l}{L/M + mask}  &  0.73   & 2.17   & 4.30\\
	 \multicolumn{2}{l}{L/M/R}  &  0.74   & 2.21   & 4.33\\
	 \multicolumn{2}{l}{L/M/R + mask}  &  \textbf{0.72}   & \textbf{2.11}  & \textbf{4.28} \\
	
\hline\hline
\end{tabularx}
\caption {\label{tab:table2} Number of Pyramids and Use of Repair Masks} 
\end{table}

\begin{table}[h!]
\centering
\begin{tabularx}{\linewidth}{ccccc}
\hline\hline

\multicolumn{2}{c}{\multirow{3}{*}{Settings}} &Chairs& \multicolumn{2}{c}{Chairs+Things} \\
 
\cline{4-5} 

     &   & Chairs&Sintel(clean)   &Sintel(final)\\
     &   & Testing& Training   & Training

	\\    \hline
	 \multicolumn{2}{l}{Level 1}    &0.73& 1.32   & 2.79\\
	 \multicolumn{2}{l}{Level 2}    &0.73& 1.33   & 2.82\\
	 \multicolumn{2}{l}{Level 3}    &0.75& 1.33   & 2.84\\
	 \multicolumn{2}{l}{Level 1+2}   &0.72& \textbf{1.24}   & \textbf{2.74}\\
	 \multicolumn{2}{l}{Level 1+2+3}   &\textbf{0.70}& 1.31   & 2.83\\
	
\hline\hline
\end{tabularx}
\caption {\label{tab:table3} Repair mask levels and numbers.} 
\end{table}

\begin{table}[h!]
\begin{tabularx}{\linewidth}{cccccc}
\hline\hline

\multicolumn{3}{c}{\multirow{3}{*}{Settings}} &\multicolumn{3}{c}{Trained on Chairs} \\
 
\cline{4-6} 

     &&   &   Chairs  & Sintel(clean)   & Sintel(final)\\
     &&   &  Testing   & Training   & Training

	\\    \hline
	 \multicolumn{3}{l}{W/M/M + mask}  &  0.81   & 2.32   & 4.39\\
	 \multicolumn{3}{l}{W/M/R + mask}  &  0.76   & 2.27   & 4.24\\
	 \multicolumn{3}{l}{W/R/R + mask}  &  \textbf{0.68}   & 2.11   & 4.30\\
	 \multicolumn{3}{l}{W/R/W + mask}  &  0.70   & \textbf{2.09}  & \textbf{4.21} \\
	
\hline\hline
\end{tabularx}
\caption {\label{tab:table4} Downsampling layer trade-off.} 
\end{table}

\subsection{Ablation Study}

Our ablation study focuses on RFPM-RAFT, because RAFT has fewer parameters but still obtains lower average end-point error (AEPE) on Sintel and KITTI-15 than RFPM-IRR-PWC. All ablation models are trained on FlyingChairs training and tested on FlyingChairs validation, and the Sintel clean and final training data. The repair mask study uses extra fine-tuning on FlyingThings3D.

We will first look at the number of pyramids. Table~\ref{tab:table2} presents the results of two or three pyramids where L, M and R express Left, Middle and Right pyramid, respectively. We find that the models benefit from the L/M/R architecture with the refinement mask, and the main reduction of the AEPE is from the refinement mask.

In Table~\ref{tab:table3}, we look at the level in the pyramids where to use a repair mask. The results indicate that the model with repair masks at both, Level 1 and Level 2 leads to the best performance in terms of AEPE. Using the repair mask at all three levels likely causes overfitting on the Chairs dataset as can be seen from low AEPE error on Chairs testing but the relatively higher AEPE on the other two datasets.  
 
Finally, we compare the different downsampling layers discussed in Section~\ref{seg:featureMap}. Table~\ref{tab:table4} shows the benefits of our module. (Note: W, M and R presents WFD, MP and RFD respectively). W/R/W shows the best performances. We suspect that RFD mainly contributes to the Sintel final pass based on the comparison between W/M/M and W/M/R. W/R/R results in the smallest testing error on Chairs, but it might be overfitting as can be seen from the higher AEPE on the Sintel final pass. 

\subsection{Transfer Learning Approach}
\label{sec:training}

\begin{table}[h!]
\centering
\begin{tabularx}{\linewidth}{ccccc}
\hline\hline

\multicolumn{1}{c}{Methods} & \multicolumn{2}{c}{Schedule} &\multicolumn{2}{c}{KITTI-15 (test)}  \\
 
\cline{4-5} 

  &&& Fl-fg(\%)  &F1-all(\%)
     
	\\    \hline
	 \multicolumn{1}{l}{RAFT}    &\multicolumn{2}{c}{C+T+S+K+H}& 6.87 & 5.10   \\
	 \multicolumn{1}{l}{RFPM-RAFT}    &\multicolumn{2}{c}{C+T+S+K+H}& \textbf{6.20} & \textbf{4.79}   \\
	 \multicolumn{1}{l}{RFPMt-RAFT}       &\multicolumn{2}{c}{$S_t$}& 6.69 & 5.08  \\

\hline\hline
\end{tabularx}
\caption {\label{tab:table6} Transfer Learning Comparison for RAFT~\cite{2020arXiv200312039T} variants on KITTI-2015. C+T+S+K+H is the training schedule used in RAFT and is also the full schedule for our RFPM-RAFT. $S_t$ is our small transfer learning schedule, which leads to our method trained as RFPMt-RAFT. The result shows that RFPMt-RAFT surpasses RAFT with a small training schedule} 
\end{table}

Transfer learning is to transfer a pre-trained representations to a novel problem which can reduce the training time of a neural network~\cite{pan2009survey}. We present a special transfer learning for our module: Given a pre-trained neural network, our goal is to improve the neural network by incorporating RFPM with only a small training schedule. We add the RFPM into pre-trained RAFT and fine-tune the overall model on KITTI-2015. We can also further improve learning by our novel Asymmetric Data Augmentation (ADA) due to the special structure of our module. ADA keeps the same geometric augmentations for Left/Middle/Right pyramids, but implements different chromatic augmentations for each batch of data. This leads us to the training schedule ($S_t$) where our module in RAFT is first trained on FlyingThings3D for 50k iteration with a batch size of 3 using Asymmetric Data Augmentation. Then we fine-tune for 100k iterations on KITTI with a batch size of 3. We call the resulting model RFPMt-RAFT to distinguish it from RFPM-RAFT that is trained with the same full C+T+S+K+H schedule as RAFT (see Section~\ref{sec:training}). Table~\ref{tab:table6} shows that our RFPMt-RAFT surpasses RAFT despite only using a small training schedule with 22.2\% iterations used by RAFT.

We also conducted an ablation study of ADA with results shown in Table~\ref{tab:table6}. MFPMt-RAFT directly trains on KITTI for 20K iteration with a batch size of three. We compare the effect of different ratios of original and augmented data presented to the RFPMt-RAFT. A probability of 0.2 (20 \%) for a sample generated with ADA appears to be most effective. Note that the numbers in Table~\ref{tab:table6} are the average of 3-fold cross validation.

\begin{table}[h!]
\centering
\begin{tabularx}{0.8\linewidth}{ccccc}
\hline\hline

\multicolumn{3}{c}{Asymmetric Data} & \multicolumn{2}{c}{KITTI-15(train)}  \\
 
\cline{4-5} 

      \multicolumn{3}{c}{Augmentation}  & F1-epe$^{+}$  &F1-all$^{+}$(\%)
     
	\\    \hline
	 \multicolumn{3}{l}{0}    &1.21& 3.63  \\
	 \multicolumn{3}{l}{0.2}    &1.19& 3.60  \\
	 \multicolumn{3}{l}{0.5}    &1.20& 3.64   \\
	 \multicolumn{3}{l}{0.8}   &1.22& 3.64  \\
	
\hline\hline
\end{tabularx}
\caption {\label{tab:table6} Asymmetric Data Augmentation (ADA) during transfer learning. $^{+}$ is the average value evaluated by 3-fold cross validation.} 
\end{table}

\section{Conclusions}
Our analysis has revealed that the traditional feature pyramid is a major reason for errors in optical flow estimation of small and finely detailed objects. The flow of these objects is lost at low resolution levels of the traditional pyramid. We propose a residual feature downsampling that includes max pooling to preserve detail features. We use multiple pyramids in our module incorporating repair masks at some levels of the pyramids. Our RFPM can be easily incorporated into modern iterative refinement optical flow methods as it only modifies the downsampling feature pyramid common to these approaches. We demonstrate that by integrating RFPM in two state-of-the-art methods, their overall error in Sintel (clean) and KITTI-15 improves but more importantly there is clear visual improvement for small and finely detailed objects. We have further proposed a special transfer learning strategy for RFPM with novel data augmentation that still achieves state-of-the-art accuracy on KITTI-2015 but with a much smaller training schedule. 

{\small
\bibliographystyle{ieee_fullname}
\bibliography{egbib,references}
}

\end{document}